\documentclass[journal]{IEEEtran}
\usepackage{amsmath,amsfonts}
\usepackage{array}
\usepackage{textcomp}
\usepackage{stfloats}
\usepackage{url}
\usepackage{verbatim}
\usepackage{graphicx}
\usepackage{cite}
\usepackage[margin=1in]{geometry}
\usepackage{xcolor}
\usepackage{pifont}
\usepackage{amssymb}
\usepackage{tabularx}
\usepackage{multirow}
\usepackage{booktabs}
\usepackage{placeins}
\usepackage{newfloat}
\usepackage{listings}
\usepackage{todonotes}
\usepackage{amsmath}
\usepackage{algorithm}
\usepackage{subcaption}
\usepackage{algorithmic}

\usepackage[margin=1in]{geometry}
\usepackage{listings}

\usepackage{xcolor}

\begin{document}

\title{DenoMAE2.0: Improving Denoising Masked Autoencoders by Classifying Local Patches}

\author{Atik Faysal,~\IEEEmembership{Student member,~IEEE,}, 
Mohammad Rostami,~\IEEEmembership{Student member,~IEEE,},
Taha Boushine,~\IEEEmembership{Student member,~IEEE,},
Reihaneh Gh. Roshan,~\IEEEmembership{Student member,~IEEE,},
Huaxia Wang,~\IEEEmembership{member,~IEEE,}
Nikhil Muralidhar,~\IEEEmembership{member,~IEEE,}
}

\markboth{Journal of \LaTeX\ Class Files,~Vol.~14, No.~8, August~2021}%
{Shell \MakeLowercase{\textit{et al.}}: A Sample Article Using IEEEtran.cls for IEEE Journals}


\maketitle

\begin{abstract}

We introduce DenoMAE2.0, an enhanced denoising masked autoencoder that integrates a local patch classification objective alongside traditional reconstruction loss to improve representation learning and robustness. Unlike conventional Masked Autoencoders (MAE), which focus solely on reconstructing missing inputs, DenoMAE2.0 introduces position-aware classification of unmasked patches, enabling the model to capture fine-grained local features while maintaining global coherence. This dual-objective approach is particularly beneficial in semi-supervised learning for wireless communication, where high noise levels and data scarcity pose significant challenges. We conduct extensive experiments on modulation signal classification across a wide range of signal-to-noise ratios (SNRs), from extremely low to moderately high conditions and in a low data regime. Our results demonstrate that DenoMAE2.0 surpasses its predecessor, DenoMAE, and other baselines in both denoising quality and downstream classification accuracy. DenoMAE2.0 achieves a 1.1\% improvement over DenoMAE on our dataset and 11.83\%, 16.55\% significant improved accuracy gains on the RadioML benchmark, over DenoMAE, for constellation diagram classification of modulation signals. 

\end{abstract}

\section{Introduction}

Semi-supervised models have emerged as a transformative paradigm in machine learning, addressing the high costs associated with data labeling \cite{zhang2024maskmatch, pmlr-v182-poulain22a}. A prevalent approach in semi-supervised learning involves pretraining models on large amounts of unlabeled data, which are abundant and readily available \cite{chen2020bigselfsupervisedmodelsstrong, xu2022revisitingpretrainingsemisupervisedlearning}. These models leverage unlabeled data to capture underlying patterns and structures, enabling effective representation learning. Consequently, the pretrained models exhibit improved performance and faster adaptation during downstream tasks, making semi-supervised learning a cost-efficient and scalable solution for real-world applications \cite{cai2022semisupervisedvisiontransformersscale}.

Recent advancements in self-supervised learning have achieved remarkable success through representation learning \cite{grill2020bootstraplatentnewapproach, chen2020exploringsimplesiameserepresentation}, contrastive learning \cite{chen2020simple, khosla2020supervised}, and masking strategies \cite{devlin2019bertpretrainingdeepbidirectional, yin2024stablemask}. Among these, masking strategies leveraging Vision Transformer (ViT) \cite{dosovitskiy2020image} architectures have set new benchmarks in tasks such as classification , reconstruction, and image analysis. Masked Autoencoders (MAE) \cite{he2021maskedautoencodersscalablevision}, in particular, have demonstrated significant potential by randomly masking portions of input data and training models to reconstruct the missing information. Despite their success, traditional MAE methods predominantly emphasize global reconstruction objectives, often neglecting fine-grained local patterns that are critical for enhancing representation learning \cite{yue2023understanding}.

Masking strategies are particularly beneficial for pretraining in wireless communication for autometic modulation classification (AMC), where acquiring labeled data is challenging due to privacy and copyright constraints \cite{zhao2024vit, zayat2023transformer}. Moreover, communication signals are often corrupted by environmental noise and channel losses, making robust representation learning essential. DenoMAE \cite{faysal2025denomae} addresses these challenges by integrating masking and denoising within a unified framework, leveraging masked modeling to enhance performance. While DenoMAE has demonstrated exceptional results in denoising and achieving strong classification accuracy, it inherits the fundamental principles of MAE, which focus primarily on global representation learning through masking. Consequently, the model remains unaware of the specific locations of the masked inputs, limiting its ability to fully exploit fine-grained local patterns and unlock the complete potential of representation learning.

We present DenoMAE2.0, an enhanced framework that introduces a novel local patch classification objective alongside the traditional reconstruction task. Our approach differs from existing methods by treating visible patches as distinct classes based on their spatial positions, enabling the model to learn position-aware local features while maintaining global coherence. This dual-objective strategy encourages the model to capture both structural and semantic information, leading to more robust and informative representations.
The key contributions of our work are threefold:

\begin{itemize}
    \item We propose a novel architecture that combines denoising reconstruction with local patch classification, enabling simultaneous learning of global and local features.
    \item We introduce a position-based classification strategy that leverages spatial information to enhance representation learning without requiring additional labels.
    \item Through extensive experiments, we demonstrate that DenoMAE2.0 achieves superior performance on various downstream and transfer learning tasks compared to traditional MAE and DenoMAE approaches, particularly in scenarios with limited labeled data.
\end{itemize}

Our results indicate that the integration of local patch classification significantly improves the quality of learned representations, leading to better denoising, generalization and faster convergence during fine-tuning. The proposed method not only advances the state-of-the-art in self-supervised learning but also provides insights into the importance of incorporating local structural information in representation learning frameworks.

\begin{figure*}[htbp]
    \centering
    \includegraphics[width=0.9\textwidth]{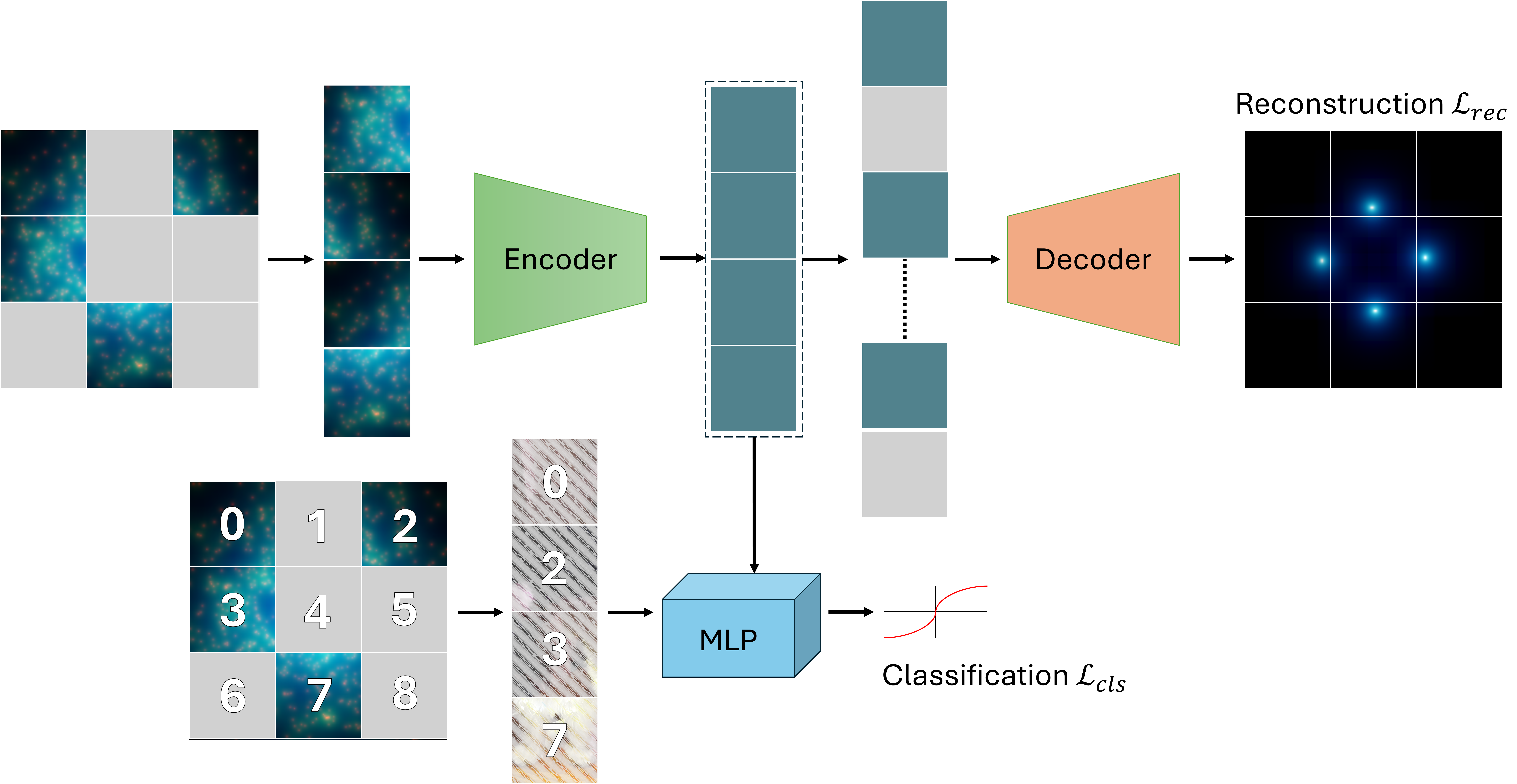}
    \caption{An example figure.}
    \label{fig:denomae2}
\end{figure*}
\section{Related Work}

\subsection{AMC in Wireless Communication}

Early AMC methods primarily relied on expert-crafted features extracted from constellation diagrams and signal statistics, utilizing likelihood-based approaches and higher-order statistics to classify modulation schemes \cite{dobre2007survey}. Constellation diagrams, which visually represent symbol distributions in the complex plane, have been a fundamental tool for feature extraction in classical AMC. These diagrams illustrate the effects of noise, interference, and channel distortions, making them valuable for differentiating modulation schemes based on clustering patterns, symbol dispersion, and trajectory structures \cite{o2018over}. 

The advent of deep learning revolutionized AMC by enabling models such as CNNs to operate directly on I/Q samples and constellation diagrams, bypassing the need for manual feature engineering \cite{zhang2021efficient}. Deep learning approaches leverage spatial patterns in constellation diagrams, learning robust representations that improve classification accuracy over conventional methods \cite{doan2020learning}. More recent advancements incorporate attention mechanisms and transformer architectures to capture long-range dependencies in signal patterns \cite{hamidi2021mcformer}. However, these deep learning-based AMC methods often require large labeled datasets and exhibit limited generalization across varying channel conditions, posing challenges for real-world deployment \cite{ya2022large}.

\subsection{Self-Supervised Learning Approaches}

The evolution of modern self-supervised learning strategies in deep learning and wireless communication has progressed through distinct phases, starting with contrastive learning, followed by representation learning, and more recently, the adoption of masking techniques.  Contrastive learning exemplified by SimCLR~\cite{chen2020simple} and MoCo \cite{he2020momentumcontrastunsupervisedvisual}, relied on distinguishing positive and negative sample pairs to learn robust feature representations, proving effective in image and signal processing tasks. This was followed by representation learning methods \cite{grill2020bootstraplatentnewapproach, berthelot2019mixmatch}, which eliminated the need for negative samples by aligning representations of augmented views, enhancing efficiency and scalability. More recently, masking-based techniques \cite{devlin2019bertpretrainingdeepbidirectional, he2021maskedautoencodersscalablevision} have gained prominence by reconstructing masked portions of inputs, offering a powerful framework for learning from incomplete data \cite{faysal2025denomae}.

In wireless communication, semi-supervised learning are particularly useful for modulation classification, where labeled data is often scarce due to the complexity of signal environments. Ma et al. \cite{ma2024refined} developed a semi-supervised framework leveraging pseudo-labeling, where a model trained on limited labeled modulation samples generates labels for unlabeled data, iteratively refining its performance across diverse modulation schemes. Another approach by Sifaou and Simeone \cite{sifaou2024semisupervisedlearningcrosspredictionpoweredinference} exploits cross-prediction-powered inference (CPPI) to enhance semi-supervised learning by mitigating the bias of synthetic labels. Their method refines ML-based predictions through tuned CPPI and meta-CPPI, achieving improved performance in tasks like beam alignment and indoor localization, particularly when labeled data is scarce. More related SSL methods in wireless communication includes convolution \cite{zhang2024sswsrnetsemisupervisedfewshotlearning, ermis2022cnn, pmlr-v77-longi17a} and transformer \cite{kong2023transformer, ren2022sigtefficientendtoendmimoofdm, kunde2023transformers} based SSLs.

\section{Methodology: DenoMAE2.0}

The overall framework of the proposed DenoMAE2.0 is illustrated in Figure \ref{fig:denomae2}. DenoMAE2.0 consists of three components: encoder, reconstruction denoising decoder, and local patch classification head. There are two complementary objectives: the denoising reconstruction objective and the local patch classification classification objective. Details are provided in Algorithm~\ref{lst:DenoMAE2.0} and introduced as follows.

\begin{lstlisting}[language=Python, caption={DenoMAE2.0, PyTorch-like pseudo code}, label={lst:DenoMAE2.0}]
# f_enc , f_dec: encoder , decoder
# f_mlp: patch classification head
# mask_r: mask ratio
# lambda_rec, lambda_class: loss function weights

for x, tgt in loader: # load a minibatch
    x = patch_emb(x) # embed patches
    x_v, x_m, mask_ids = masking(x, mask_r) # random split visible and masked patches
    q_v = f_enc(x_v) # local patch features
    logits = f_mlp(q_v) # predicted labels
    k_m = f_dec(q_v) # reconstructed pixels
    loss = lambda_rec * lambda_cls(k_m, x_m) + lambda_cls * cls_loss(logits, tgt)
    loss.backward()
    update() # optimizer update

def class_labels(mask_ids):
    classes = ((img_size//patch_size)**2) * (1 - mask_r)
    labels = classes[mask, where mask == 0] # 0: visible patches, 1: masked patches
    return labels

def rec_loss(k, x): # reconstruction
    x = norm_pix(x) # normalize every patch
    loss = (k - x) ** 2 # compute MSE loss over masked patches
    return loss

def cls_loss(logits, tgt): # classification
    loss = CrossEntropyLoss(logits, tgt)
    return loss
\end{lstlisting}

\subsection{Image Masking and Encoding}

The preprocessing stage transforms an input image $\mathbf{x} \in \mathbb{R}^{H \times W \times C}$ into a sequence of non-overlapping patches $\mathbf{x}_p \in \mathbb{R}^{N \times (P^2 \cdot C)}$, where $(H, W)$ defines the spatial dimensions, $C$ denotes the number of channels, and $(P, P)$ specifies the patch size. The total number of patches is $N = HW/P^2$. 

These patches undergo linear projection through a PatchEmbed layer to obtain initial embeddings. We adopt a masking strategy that randomly removes a substantial portion (75\%) of the patches. Let $\mathbf{x}_v$ and $\mathbf{x}_m$ represent the visible and masked patches respectively. The visible patches $\mathbf{x}_v$ are combined with learned positional embeddings and processed by a Vision Transformer (ViT) encoder to generate patch-level features $\mathbf{q}_v$. These features form the basis for downstream reconstruction and classification tasks.

\subsection{Decoder Architecture and Reconstruction}

Since the length of visible patch features $\mathbf{q}_v$ is smaller than the total number of image patches $N$, we augment $\mathbf{q}_v$ with mask tokens to construct a complete feature set. Each mask token is a shared, learnable vector that indicates positions requiring reconstruction. Following the encoding process, we enhance this complete feature set with positional embeddings before processing it through a transformer-based decoder. The decoder's output undergoes transformation through a linear projection layer (not depicted in Figure 1) to map features back to pixel space. 

The reconstruction objective $\mathcal{L}_{\text{rec}}$ employs mean squared error (MSE) between the reconstructed and original images. Consistent with established approaches, we compute this loss exclusively on the masked patches:

\begin{equation}
    \mathcal{L}_{\text{rec}} = \text{MSE}(\hat{\mathbf{x}}_m, \mathbf{x}_m)
\end{equation}

where $\hat{\mathbf{x}}_m$ represents the reconstructed patches and $\mathbf{x}_m$ denotes the original masked patches.

\subsection{Classification Branch}

The visible patch features $\mathbf{q}_v$ are additionally leveraged for classification. Diverging from traditional classification, where one input contains a single class, we assign visible patches to classes based on their spatial positions. The classification branch takes only the class tokens from the encoder to classify them. The total number of classes is the same as the number of visible patches which is calculated as:

\begin{equation}
    N_{\text{classes}} = \left(\frac{\text{img\_size}}{\text{patch\_size}}\right)^2 \cdot (1 - r_{\text{mask}})
\end{equation}

Since we obtain several classes from a single image, the classification branch is a multi-class classification problem. The classification head consists of a simple linear layer that projects the patch features directly to the number of classes. The classification loss $\mathcal{L}_{\text{cls}}$ employs cross-entropy (CE) between predicted and ground truth labels:

\begin{equation}
    \mathcal{L}_{\text{cls}} = \text{CE}(\mathbf{p}, \mathbf{y})
\end{equation}

where $\mathbf{p}$ denotes the predicted probabilities and $\mathbf{y}$ represents the ground truth labels.

\subsection{Overall Objective}

Our DenoMAE2.0 optimizes a combination of reconstruction and classification losses, enabling simultaneous learning of fine-grained local and global features. The overall loss function is formulated as a weighted sum:

\begin{equation}
    \mathcal{L} = \lambda_{\text{rec}}\mathcal{L}_{\text{rec}} + \lambda_{\text{cls}}\mathcal{L}_{\text{cls}}
\end{equation}

where $\lambda_{\text{rec}}$ and $\lambda_{\text{cls}}$ are balancing weights for the reconstruction and classification objectives, respectively. Through empirical validation (see Table~\ref{tab:weights_accuracy}), we set $\lambda_{\text{rec}} = 1.0$ and $\lambda_{\text{cls}} = 0.1$.

\subsection{Fine-Tuning on Downstream Tasks}

Following pre-training, the DenoMAE2.0 encoder is fine-tuned on specific recognition tasks to enhance performance. During this stage, the decoder and classification head are excluded, retaining only the encoder. For fine-tuning, the model utilizes the complete set of patches, corresponding to the uncorrupted input images, to optimize for downstream recognition tasks.

\section{Datasets}

\subsection{Constellation Diagram Generation}

Constellation diagrams offer a more informative approach to modulation classification compared to raw time-series signals, capturing richer detail essential for accurate interpretation \cite{doan2020learning}. The process of generating constellation diagrams for modulation signal classification involves multiple stages. We begin by mapping modulated signals onto a $7 \times 7$ complex plane, which provides sufficient space to capture signal samples while maintaining computational efficiency for SNR ranges of -10 dB to 10 dB. The basic constellation diagram is then enhanced through a multi-step process: first, as a gray image that handles varying pixel densities where multiple samples may occupy single pixels; second, as an enhanced grayscale image that employs an exponential decay model ($B_{i,j}$) to account for both the precise position of samples within pixels and their influence on neighboring pixels. This model considers the sample point's power ($P$), the distance between sample points and pixel centroids ($d_{i,j}$), and an exponential decay rate ($\alpha$) \cite{peng2018modulation}. To make the representation compatible with DenoMAE2.0, which expects RGB input, we generate a three-channel image by creating three distinct enhanced grayscale images from the same data samples, each utilizing different exponential decay rates. 

The pretraining dataset comprises 10,000 samples uniformly distributed across ten modulation types, with randomly assigned SNR values in the range of -10 dB to 10 dB. For the downstream classification task, we use 1000 samples for training and 100 samples for testing, evenly distributed among the 10 classes, evaluated at SNR values from -10 dB to 10 dB in 1 dB increments.

\subsection{Parameters in Sample Generation} 

The dataset comprises signals from ten modulation formats (see appendix, Table~\ref{tab:cons_classes}, for detailed modulation types), each originally of length $L_0 = 1024$. To conform to the transformer's input dimensions, signals undergo a two-step preprocessing: (1) reshaping to $\mathbf{S}_1 \in \mathbb{R}^{32 \times 32}$, and (2) interpolation to $\mathbf{S}2 \in \mathbb{R}^{224 \times 224}$. For formats with $L_0 > 1024$ (e.g., $L_\text{GMSK} = 8196$), signals are initially downsampled to $L_0$. To match the three-channel structure of constellation images $\mathbf{I} \in \mathbb{R}^{3 \times 224 \times 224}$, $\mathbf{S}_2$ is replicated across three channels, yielding $\mathbf{S}_3 \in \mathbb{R}^{3 \times 224 \times 224}$. All signals are sampled at $f_s = 200$ kHz, ensuring consistency across modulation formats. 

\section{Experimental Results}

\subsection{Denoising Performance}

\begin{figure*}[htbp]
    \centering
    \includegraphics[width=0.7\linewidth]{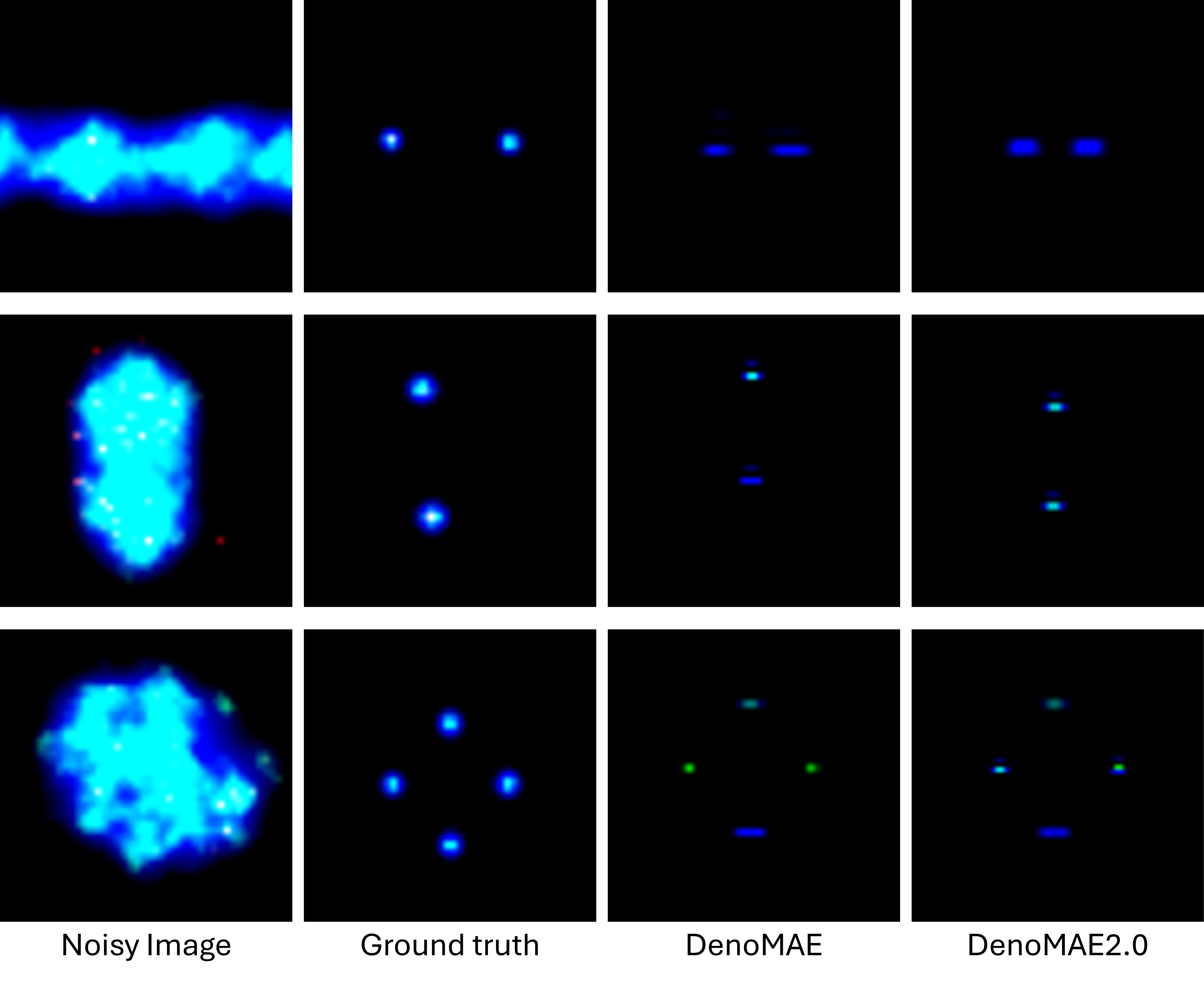}
    \caption{Reconstruction performance comparison between DenoMAE and DenoMAE2.0}
    \label{fig:recon}
\end{figure*}

During the pretraing DenoMAE2.0 learns to denoise and reconstruct the missing patches and thereby learn their inherent representation. Therefore, we evaluate this denoisng and reconstruction performance during and after the pretraining phase. The effectiveness of DenoMAE2.0 is evident in the reconstructed constellation diagrams shown in Figure \ref{fig:recon}. Compared to DenoMAE, DenoMAE2.0 achieves superior noise reduction while preserving key signal features crucial for modulation classification. Table \ref{tab:ssim_psnr_comparison} quantifies this improvement using two standard metrics: structural similarity index (SSIM) and peak signal-to-noise ratio (PSNR). Interestingly, for the first image, DenoMAE achieved higher SNR and PSNR despite DenoMAE2.0 producing a visually superior reconstruction. However, for the remaining two samples, DenoMAE2.0 outperformed DenoMAE in both SSIM and PSNR.

\begin{table}[htbp]
    \centering
    \caption{Comparison of SSIM and PSNR for DenoMAE and DenoMAE2.0}
    \renewcommand{\arraystretch}{1.2}
    \begin{tabular}{|c|c|c|c|}
        \hline
        Method & Image & DenoMAE & DenoMAE2.0 \\ \hline
        \multirow{3}{*}{SSIM} 
        & Image1 & 0.9660 & 0.9634 \\ 
        & Image2 & 0.9407 & 0.9607 \\ 
        & Image3 & 0.9422 & 0.9466 \\ \hline
        \multirow{3}{*}{PSNR (dB)} 
        & Image1 &  42.9054 & 42.3228 \\ 
        & Image2 & 42.0005 & 42.1061 \\ 
        & Image3 & 40.3067 & 40.7508 \\ \hline
    \end{tabular}
    \label{tab:ssim_psnr_comparison}
\end{table}

\subsection{Latent Space Visualization with TSNE}

\begin{figure*}
    \centering
    \subfloat{%
       \includegraphics[trim=1cm 1cm 1cm 1cm,clip,
       width=0.25\linewidth]{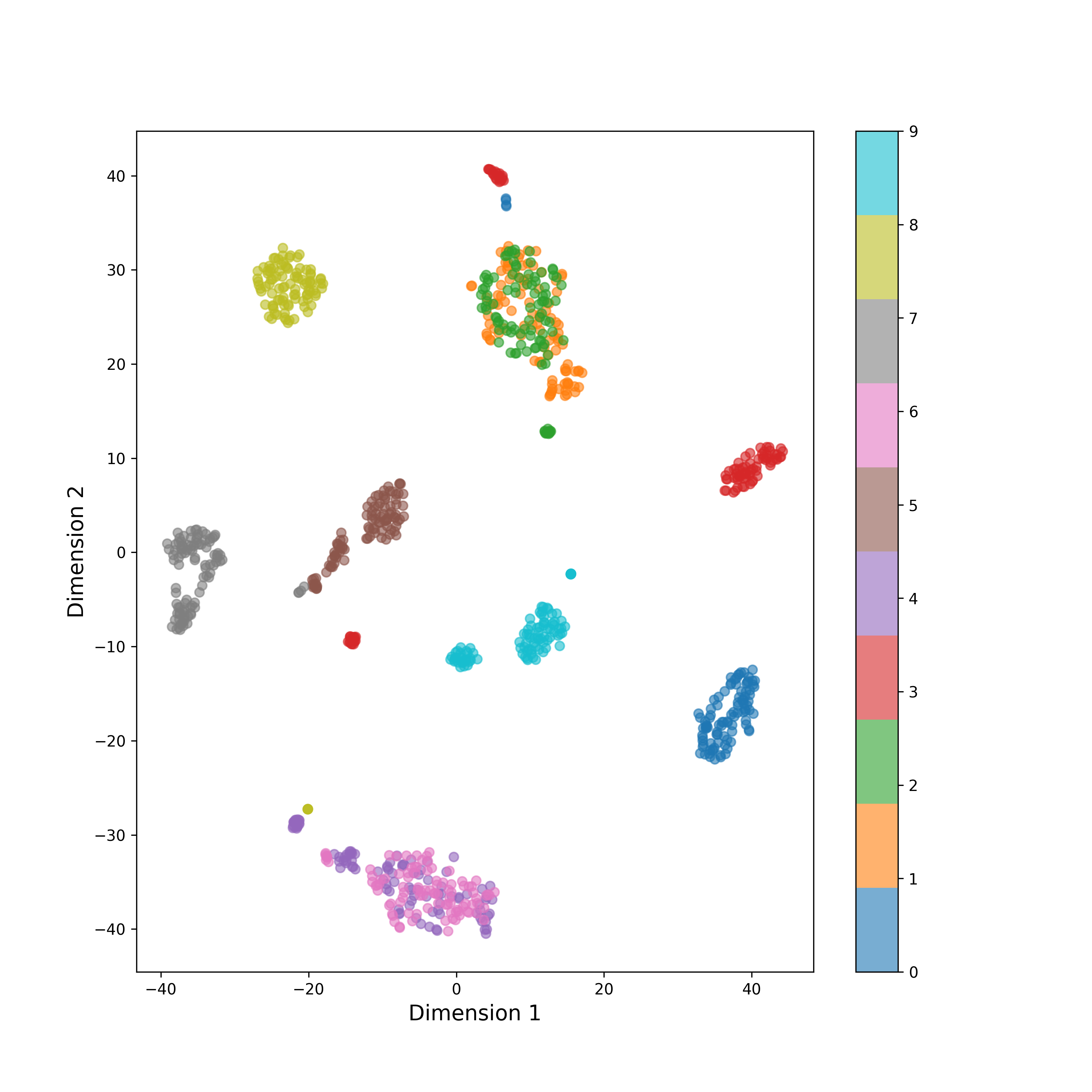}}
    \hfill
    \subfloat{%
        \includegraphics[trim=1cm 1cm 1cm 1cm,clip,
        width=0.25\linewidth]{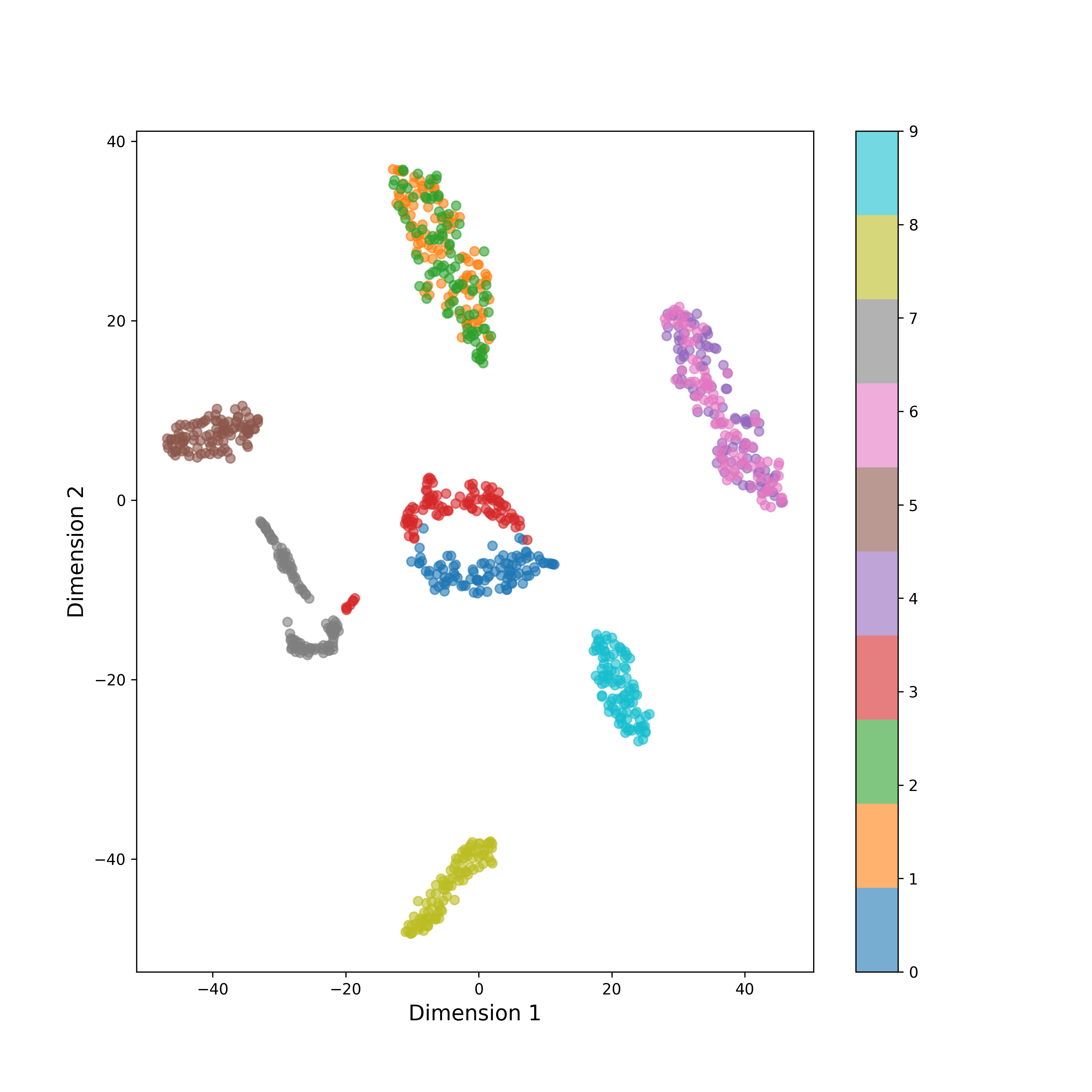}}
    \hfill
    \subfloat{%
        \includegraphics[trim=1cm 1cm 1cm 1cm,clip,
        width=0.25\linewidth]{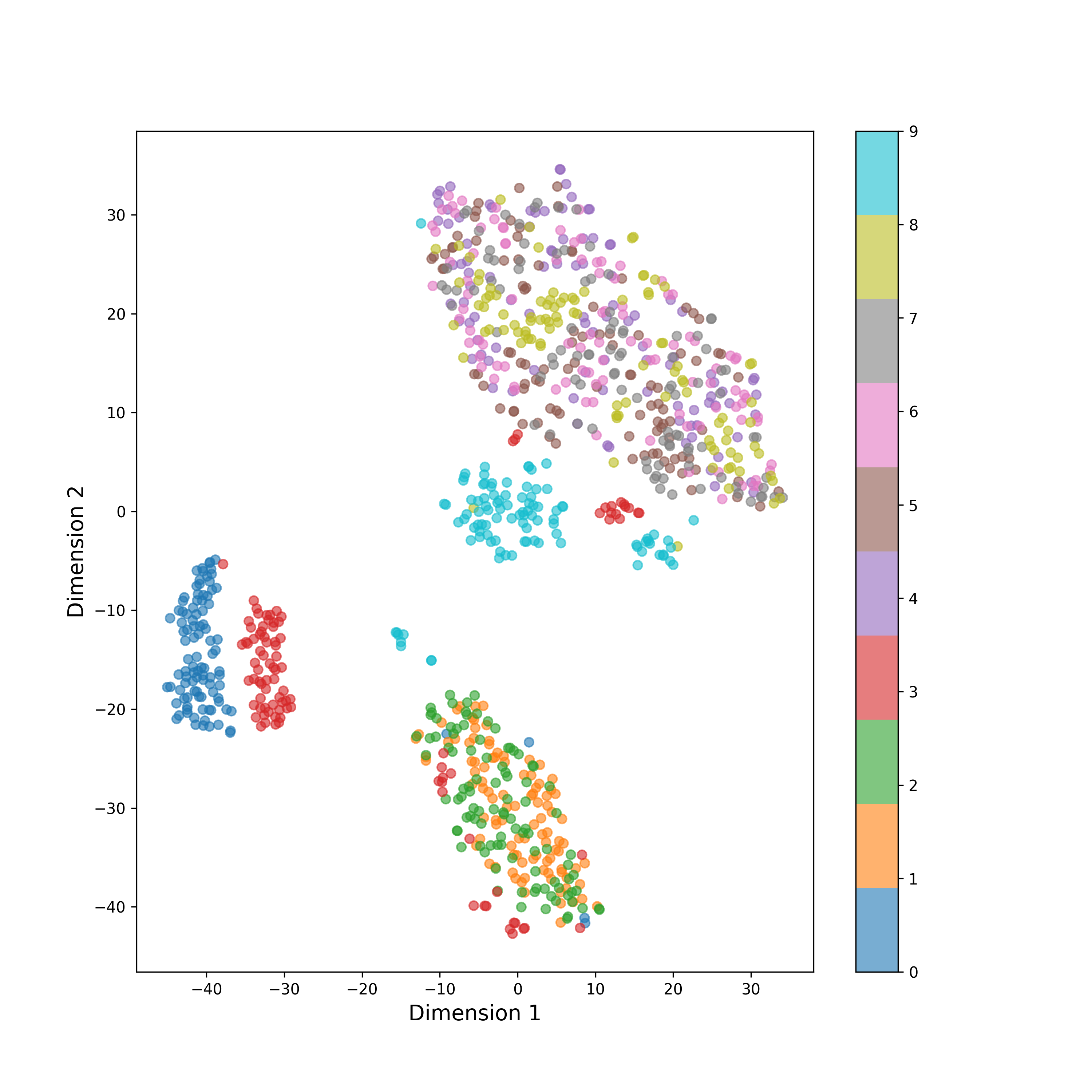}}
    \hfill
    \subfloat{%
        \includegraphics[trim=1cm 1cm 1cm 1cm,clip,
        width=0.25\linewidth]{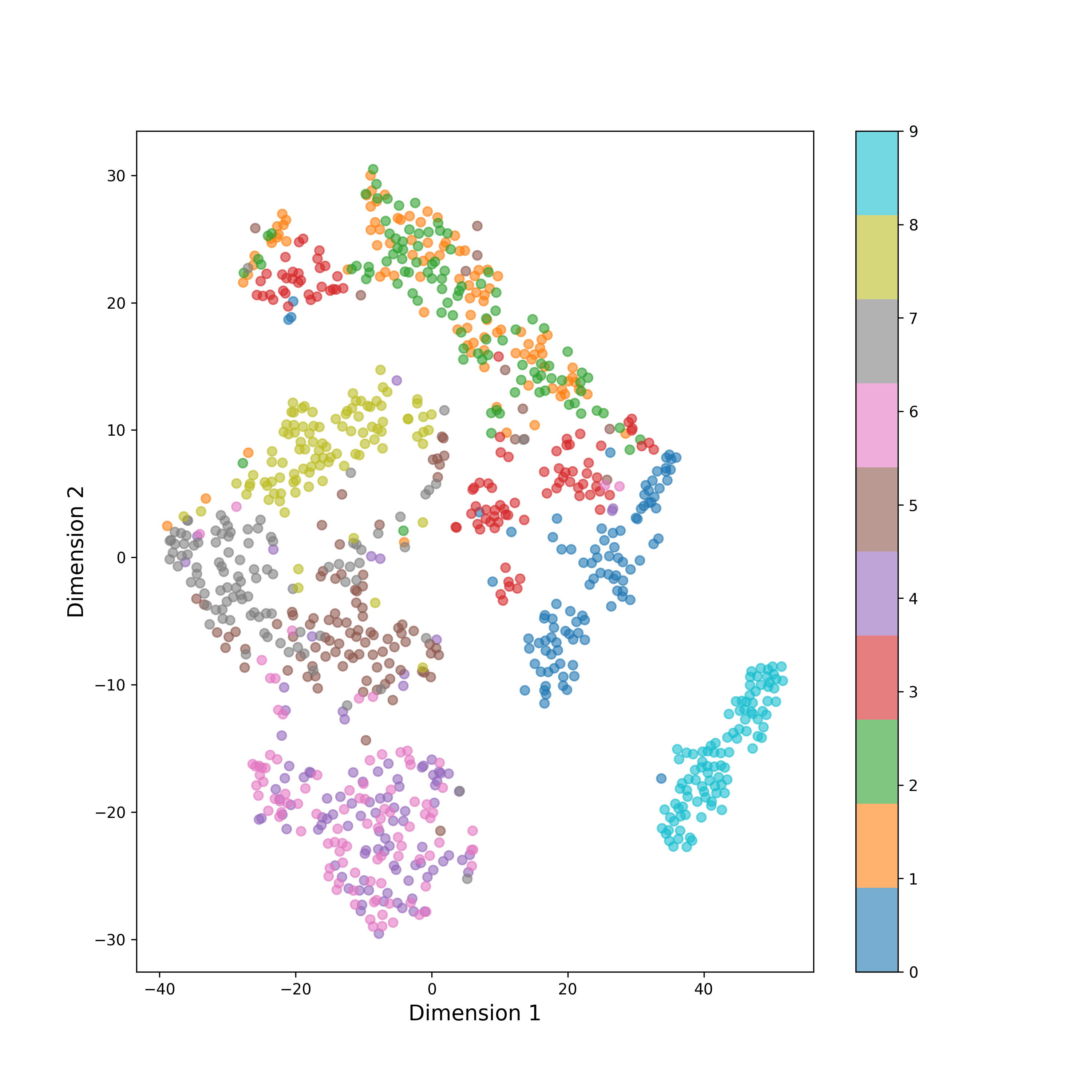}}
    \caption{Latent representation visualization using t-SNE. From left to right: (1) DenoMAE without masking, (2) DenoMAE2.0 without masking, (3) DenoMAE with 0.75\% masking, and (4) DenoMAE2.0 with 0.75\% masking.}
    \label{fig:tsne} 
\end{figure*}

In Figure~\ref{fig:tsne}, we visualize the latent representations of DenoMAE and DenoMAE2.0 using t-distributed stochastic neighbor embedding (t-SNE)~\cite{van2008visualizing}. The plots (left two) reveal that DenoMAE2.0 produces more accurate and distinct clusters than DenoMAE, indicating superior feature separation and representation learning. Specifically, in DenoMAE, classes 0, 3, and 9 contain more separated clusters within the same class. In contrast, DenoMAE2.0 exhibits more distinct clusters for these classes.

This enhanced clustering is particularly evident in the masked samples (right two plots), where we mask out a significant portion (75\%) of the images. Masking typically makes distinct cluster formation more challenging. However, DenoMAE2.0 achieves clearer boundaries between classes, suggesting improved robustness and generalization capabilities. In DenoMAE, classes 0, 3, and 9 achieve somewhat distinct clusters, with a few samples overlapping in other clusters. In DenoMAE2.0, classes 0, 3, 6, 7, 8, and 9 show better-separated clusters, indicating enhanced feature learning and representation.

However, in all the plots, classes 1 and 2 entirely overlap with each other, indicating that the model struggles to distinguish between these two classes. This is a common issue in modulation classification, as some classes are very similar to each other.

\subsection{Finetuning Accuracy}

After pretraining, we transfer the learned representations to the downstream classification task and compare DenoMAE2.0 with state-of-the-art models. As shown in Table \ref{tab:example}, when finetuned on the downstream task, DenoMAE2.0 achieves the highest test accuracy of 82.40\% among all compared methods. This represents a significant improvement over the baseline ViT (79.90\%) and other representation learning approaches including DEiT (81.20\%), MoCov3 (81.00\%), BEiT (80.40\%), MAE (80.10\%), and the original DenoMAE (81.30\%). The consistent performance advantage demonstrates that DenoMAE2.0's enhanced denoising strategy learns more effective representations during pretraining that transfers well to the downstream classification task.

\begin{table}[htbp]
    \centering
    \caption{Downstream classification accuracy for different models}
    \resizebox{0.8\linewidth}{!}{%
    \begin{tabular}{|c|c|}
        \hline
        Method & Test accuracy (\%) \\ \hline
        ViT & 79.90 \\ \hline
        DEiT & 81.20 \\ \hline
        MoCov3 & 81.00 \\ \hline
        BEiT & 80.40 \\ \hline
        MAE &  80.10 \\ \hline
        DenoMAE &  81.30 \\ \hline
        DenoMAE2.0 &  \textbf{82.40} \\ \hline
    \end{tabular}%
    }
    \label{tab:example}
\end{table}

\begin{figure}[htbp]
    \centering
    \includegraphics[width=0.9\linewidth]{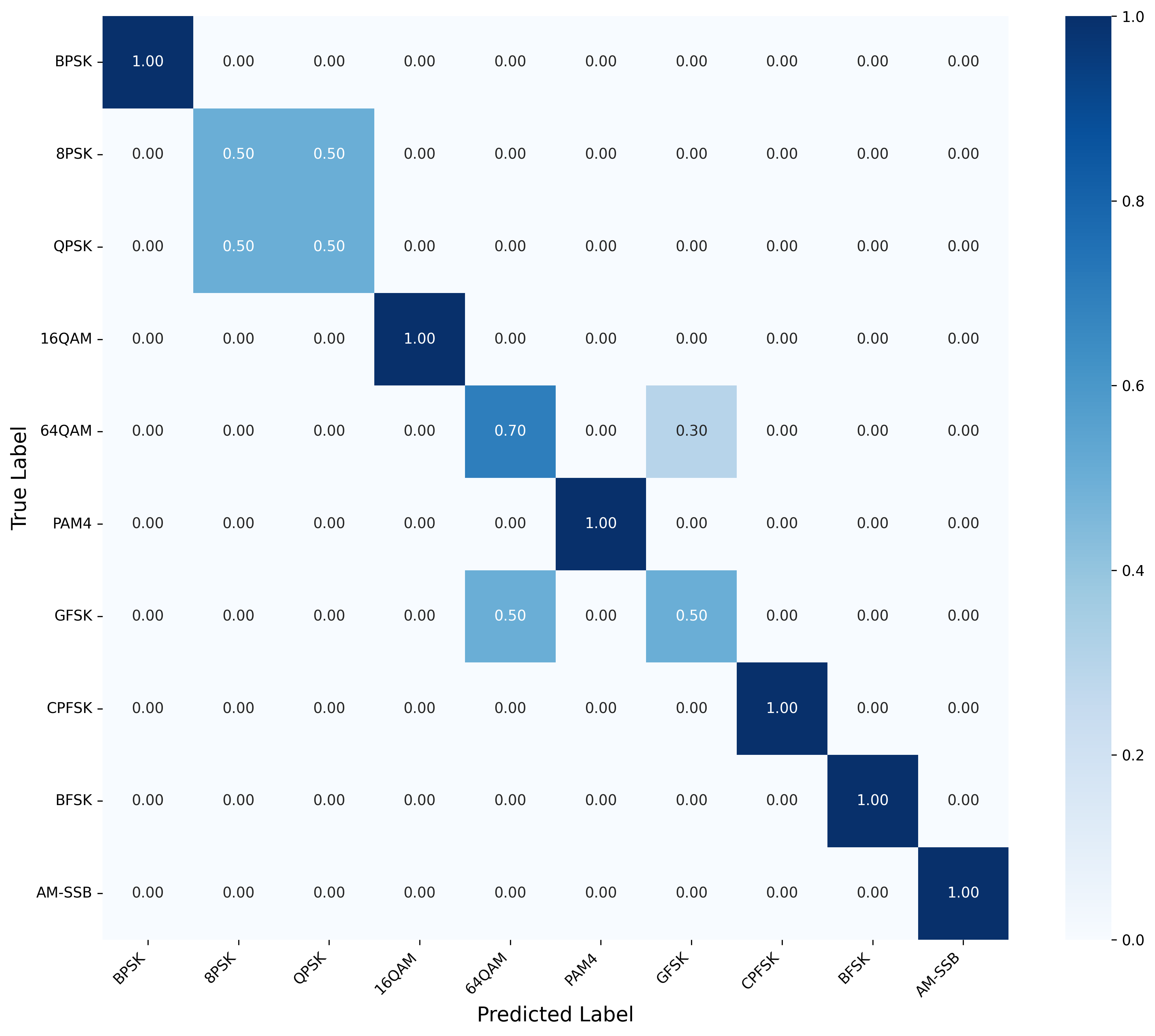}
    \caption{Confusion matrix for DenoMAE2.0 downstream classification} 
    \label{fig:confusion}
\end{figure}

Figure~\ref{fig:confusion} presents a confusion matrix visualizing the per-class performance of DenoMAE2.0. The diagonal elements indicate correct classifications, while off-diagonal elements show misclassifications. The matrix reveals strong performance across most classes, though notable confusion exists between classes 1 and 2, aligning with observations from the t-SNE analysis. This confusion pattern suggests inherent similarities in the signal characteristics of these two modulation types that challenge even our enhanced model architecture.

\subsection{Performance on Number of Epochs}

\begin{figure}[htbp]
    \centering
    \includegraphics[width=0.9\linewidth]{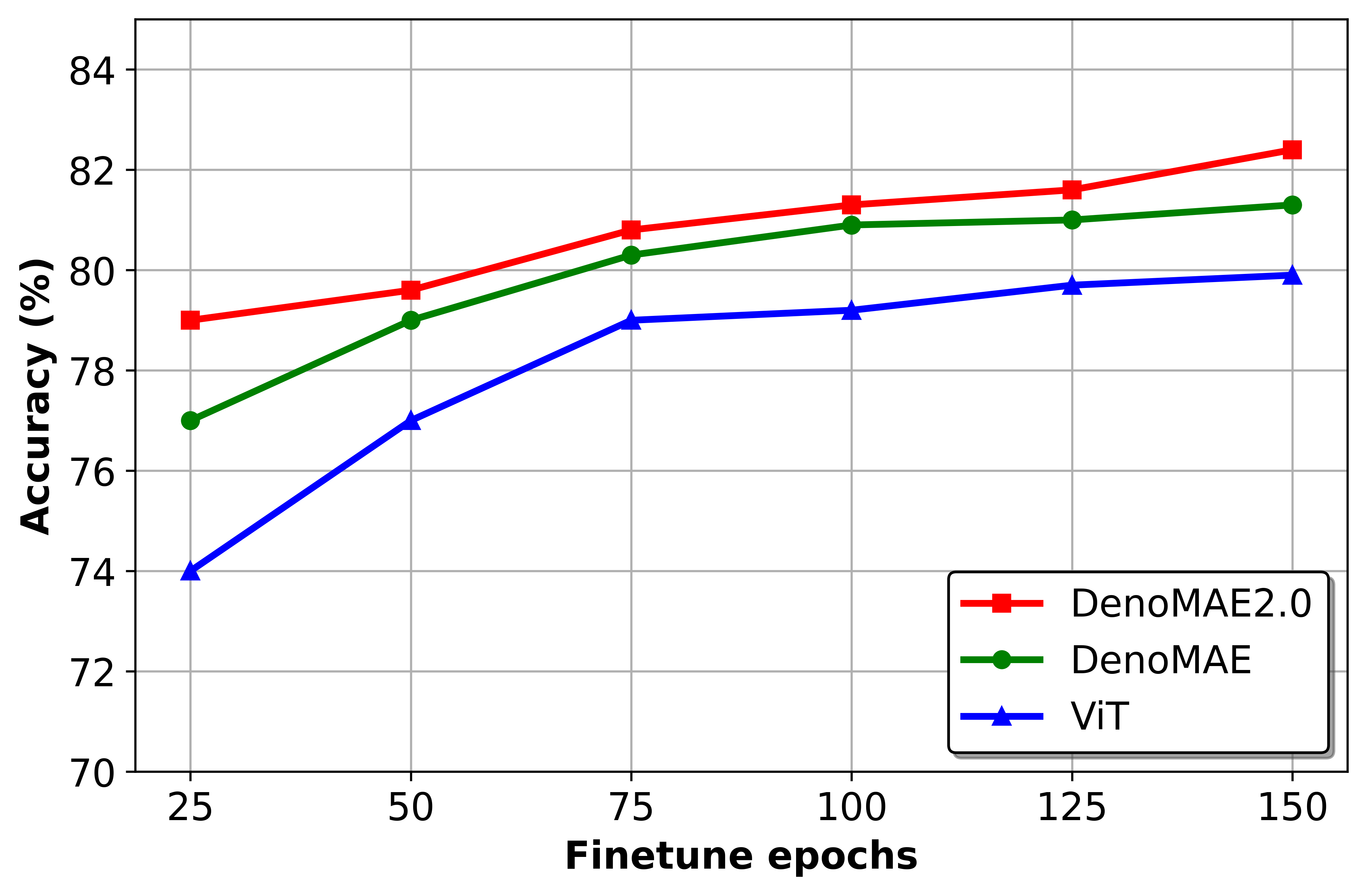}
    \caption{Epoch finetuning.}
    \label{fig:epoch}
\end{figure}

Figure~\ref{fig:epoch} illustrates the finetuning performance across different epochs. The plot shows a steady improvement in model accuracy as the number of epochs increases, with DenoMAE2.0 consistently outperforming baseline methods. DenoMAE achieves higher accuracy than ViT continuously. On the other hand, although the gap between DenoMAE and ViT is larger than DenoMAE and DenoMAE2.0, DenoMAE2.0 always obtains better accuracy than DenoMAE.

\subsection{Performance Across Different SNRs}

The SNR performance plot demonstrates robust classification accuracy across various SNRs ranging from -10 dB to 10 dB. As the SNR values increases all the methods perform better and when the SNR is low the performance degrades. ViT, which has no pertaining phase, presumably obntains the lowest performance in all cases. MAE, which pretrains on modulation signals but has no denoising capability, obtains a slight higher accuracy than ViT. DenoMAE obtains higher accuracy than MAE in most cases however for SNE of 2 and 3 we observe that MAE obtains a higher accuracy. On the other hand, DenoMAE2.0 outperforms all the methods for all the SNRs.  Noteably, all the methods lose significant accuracy for snr lower than -2 dB. However, DenoMAE2.0 maintains comparatively robust performance down to -7 dB, demonstrating its enhanced resilience to noise.

\begin{figure}[htbp]
    \centering
    \includegraphics[width=\linewidth]{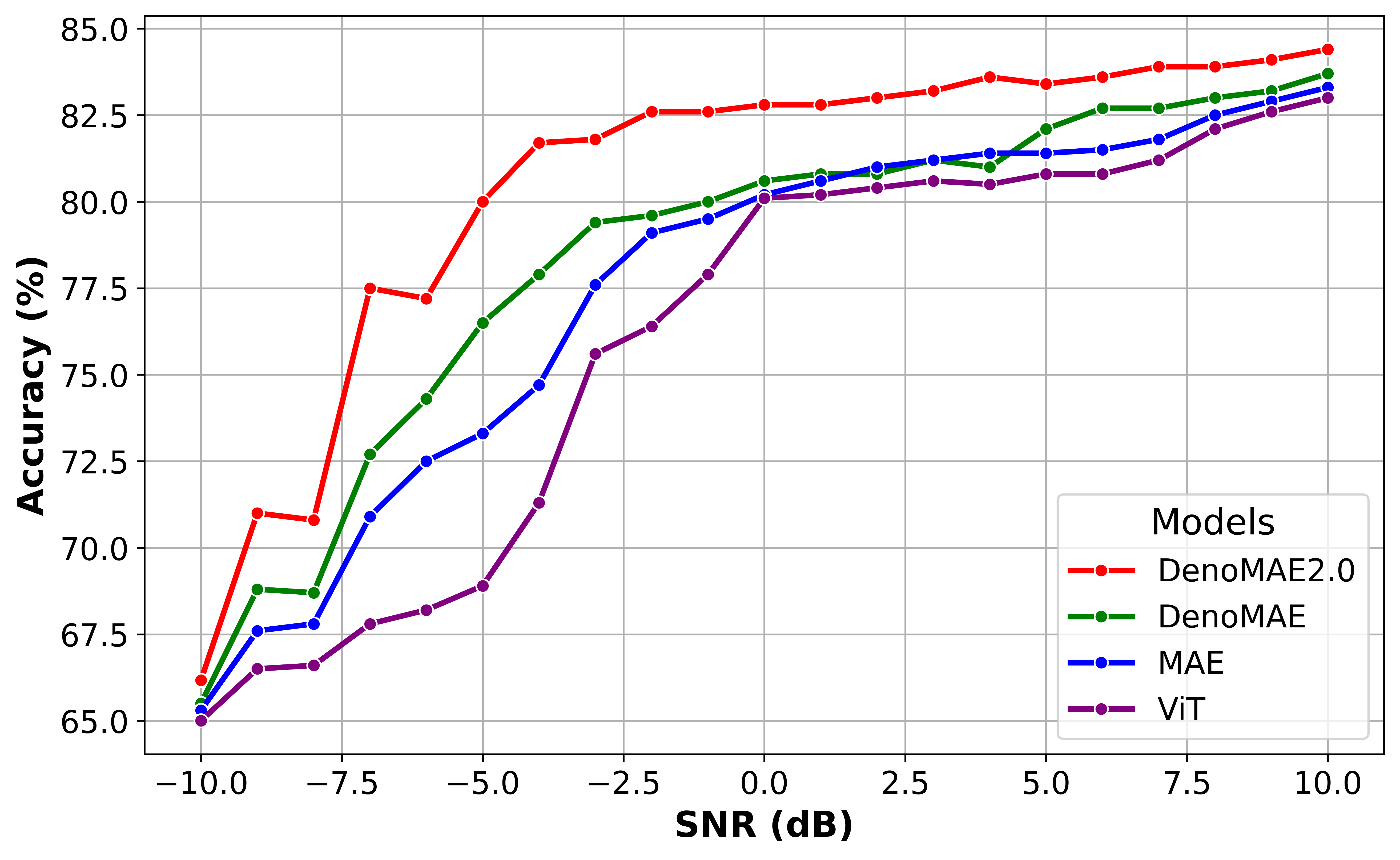}
    \caption{Accuracy comparison across different SNRs}
    \label{fig:snr_acc}
\end{figure}

\section{Comparison with Other AMC Methods}

Table~\ref{tab:compare} presents a comprehensive comparison between DenoMAE2.0 and other state-of-the-art modulation classification methods. AlexNet achieved 82.8\% accuracy with 8 classes at 2dB SNR using a large dataset of 800,000 samples. NMformer reported 71.6\% accuracy across 10 classes with SNR ranging from 0.5 to 4.5dB using 106,800 samples. CNN-AMC and DL-GRF showed relatively lower performance with 50.40\% and 59\% accuracy respectively, though they were tested under different conditions. The original DenoMAE achieved 81.3\% accuracy across 10 classes with SNR ranging from 0 to 10dB. Our proposed DenoMAE2.0 demonstrates superior performance with 82.4\% accuracy under similar conditions, using only 10,000 samples for pretraining and 1,000 samples for fine-tuning, indicating improved efficiency in both performance and data utilization.

\begin{table*}[htbp]
    \centering
    \caption{Comparison of DenoMAE2.0 with other modulation classification methods}
    \label{tab:compare}
    \begin{tabular}{cccccc}
    \toprule
    \multirow{2}{*}{\textbf{Methods}} & \multicolumn{2}{c}{\textbf{Number of samples}} & \multirow{2}{*}{\textbf{SNR dB}} & \multirow{2}{*}{\textbf{Number of classes}} & \multirow{2}{*}{\textbf{Test accuracy (\%)}} \\ 
    \cmidrule(lr){2-3}
     & \textbf{Pretraining} & \textbf{Fine-tuning} & & \\ \midrule
    
    AlexNet \cite{peng2018modulation}  & 800,000 (train) & 8,000 (test)  & 2 & 8 & 82.8  \\
    NMformer \cite{faysal2024nmformer} & 106,800 & 3,000 & 0.5 - 4.5 & 10 & 71.6 \\
    CNN-AMC \cite{meng2018automatic} & 79,200 & 228,060 & -6 & 4 & 50.40 \\
    DL-GRF \cite{sun2022automatic} &  2,000 (train) & 200 (test) & 0 & 4 & 59 \\
    DenoMAE \cite{faysal2025denomae} & 10,000 & 1,000 & 0 - 10 & 10 & 81.3 \\
    DenoMAE2.0 (\textbf{Ours}) & 10,000 & 1,000 & 0 - 10 & 10 & 82.4 \\
    
    \bottomrule
    \end{tabular}
\end{table*}

\subsection{Transferability}

To show the transferability of DenoMAE2.0 on other similar task, we performed experiments on a similar benchmark dataset named RadioML. A brief description on the RadioML dataset is provided in the following subsection.

\subsubsection{RadioML Dataset}

RadioML 2018.01A~\cite{o2018over} is a large-scale dataset designed for AMC and machine learning-based RF signal processing. It consists of 2,555,904 IQ (in-phase and quadrature) signal samples, where each sample is represented as a 1024×2 array, capturing both in-phase and quadrature components. The dataset spans 24 modulation types, including digital and analog schemes such as BPSK, QPSK, 8-PSK, 16-QAM, AM, and FM, making it a comprehensive resource for signal classification tasks. Signals are generated under 26 distinct signal-to-noise ratio (SNR) levels, ranging from -20 dB to +30 dB in 2 dB increments, simulating real-world conditions affected by fading, interference, and channel noise. Each modulation type and SNR combination contains 4096 samples, ensuring a balanced distribution for model training and evaluation. The dataset, widely used as a benchmark in wireless communication~\cite{ccamlibel2024automatic} and spectrum sensing research~\cite{elyousseph2021deep}, facilitates advancements in deep learning-based modulation recognition, interference mitigation, and cognitive radio applications~\cite{cheng2024automatic, jagatheesaperumal2024deep, 10014805}.

To keep a consistent same scale implementation, we used 100 samples in each class for the 24 classes in RadioML for training and 10 samples in each class for testing. Therefore, we used only 2400 samples for training and 240 samples for testing. As a result we only used 0.103\% of the total dataset. We only obtain SNRs of -20, -10, 0, 10, 20 dB to show performance across low and high SNRs, as well as interpolaribility. 

\subsubsection{Performance}

\begin{figure}[htbp]
    \centering
    \includegraphics[width=\linewidth]{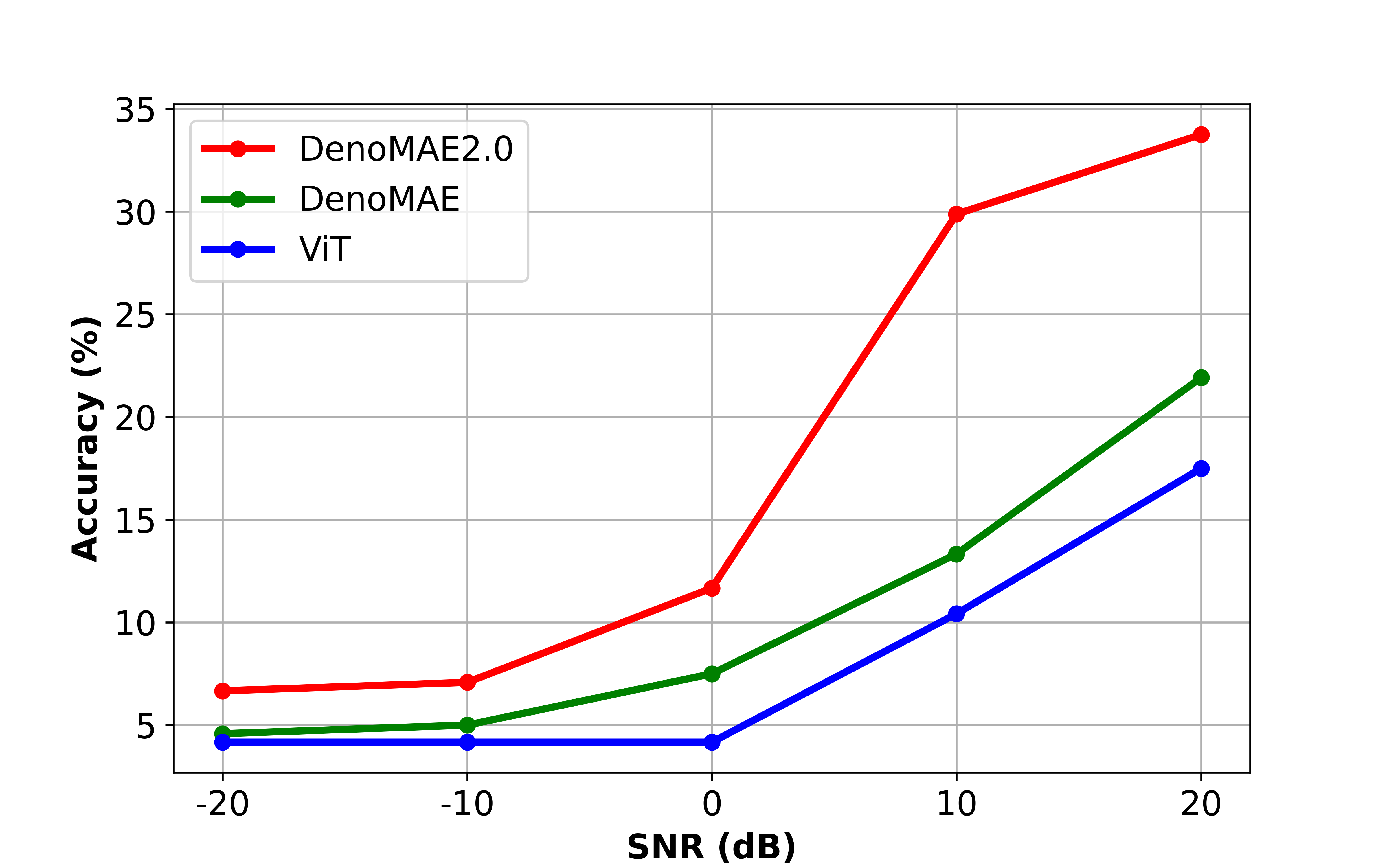}
    \caption{Transfer learning performance on RadioML dataset}
    \label{fig:trans}
\end{figure}

The transfer learning results are shown in Figure~\ref{fig:trans}. Performance improves with increasing SNR, with DenoMAE2.0 consistently outperforming baseline methods across all SNR values. At 20 dB SNR, DenoMAE2.0 achieves 33.75\% accuracy compared to 21.92\% for DenoMAE, representing an 11.83\% improvement. Similarly at 10 dB SNR, DenoMAE2.0 obtains 29.88\% accuracy versus 13.33\% for DenoMAE, a 16.55\% gain. For out-of-distribution SNRs between -10 dB and -20 dB, ViT performs at chance level (4.27\%) while DenoMAE shows minimal improvement. However, DenoMAE2.0 maintains significant accuracy of 7.08\% at -10 dB and 6.67\% at -20 dB, demonstrating enhanced transferability and robustness.

\section{Ablation Study}

In this section, we analyze the impact of various components on the performance of our model.

\subsection{Effect of the Loss Function}

We conducted experiments to evaluate the contribution of different loss components in our model. As shown in Table~\ref{tab:loss}, we analyzed three configurations: using only reconstruction loss, only classification loss, and combining both losses. The results demonstrate that while reconstruction loss alone achieves reasonable performance (81.30\%), classification loss by itself underperforms (69.80\%). However, combining both losses yields the best results (82.40\%), suggesting that the joint optimization of reconstruction and classification objectives leads to more robust feature learning and better overall performance.

\begin{table}[htbp]
    \centering
    \caption{Accuracy for different loss configurations}
    \begin{tabular}{ccc}
        \toprule
        Reconstruction Loss & Classification Loss & Accuracy (\%) \\
        \midrule
        \ding{51} &     & 81.30 \\
                 & \ding{51} & 69.80 \\
        \ding{51} & \ding{51} & 82.40 \\
        \bottomrule
    \end{tabular}
    \label{tab:loss}
\end{table}

\subsection{Effect of Auxiliary Loss Weights}

We conducted experiments to analyze the impact of auxiliary loss weights on model performance, as shown in Table~\ref{tab:weights_accuracy}. The results demonstrate that the model's accuracy varies with different weight values, with 0.10 achieving the optimal performance at 82.4\%. We observed a clear trend where very high weights (1.0) significantly degraded performance to 79.60\%, while moderate weights (0.25, 0.50) maintained consistent performance at 82.0\%. Lower weights (0.01, 0.05) showed slightly reduced performance at 81.3\% and 81.5\% respectively, indicating that the auxiliary loss weight needs to be carefully tuned to achieve the best balance between learning objectives.

\begin{table}[htbp]
    \centering
    \caption{Accuracy for different auxiliary loss weights}
    \resizebox{\columnwidth}{!}{%
    \begin{tabular}{ccccccc}
        \toprule
        Aux. loss weights & 0.01 & 0.05 & 0.10 & 0.25 & 0.50 & 1.0 \\
        \midrule
        Accuracy (\%) & 81.3 & 81.5 & 82.4 & 82.0 & 82.0 & 79.60 \\
        \bottomrule
    \end{tabular}%
    }
    \label{tab:weights_accuracy}
\end{table}

\subsection{Combination of Losses}

We investigated the effect of different weight combinations between reconstruction and auxiliary losses, as shown in Table~\ref{tab:weights_comb}. The results reveal that balancing these two losses is crucial for optimal performance. Equal weighting (0.5, 0.5) yields suboptimal performance at 78.80\%, while gradually increasing the reconstruction loss weight and decreasing the auxiliary loss weight shows steady improvement. The optimal combination is achieved with reconstruction loss weight of 1.0 and auxiliary loss weight of 0.10, resulting in 82.40\% accuracy. Further increasing the reconstruction loss weight to 1.2 while reducing auxiliary loss weight to 0.05 leads to a slight performance degradation (82.0\%), suggesting that maintaining a proper balance between these losses is essential for optimal model performance.

\begin{table}[htbp]
    \centering
    \caption{Accuracy for different combination of reconstruction and auxiliary loss weights}
    \label{tab:weights_comb}
    \resizebox{\columnwidth}{!}{%
    \begin{tabular}{ccccccc}
        \toprule
        Rec. loss Weights & 0.5 & 0.75 & 0.80 & 0.90 & 1.0 & 1.2 \\
        \midrule
        Aux. loss weights & 0.5 & 0.25 & 0.10 & 0.10 & 0.10 & 0.05 \\
        \midrule
        Accuracy (\%) & 78.80 & 80.50 & 81.60 & 82.0 & 82.40 & 82.0 \\
        \bottomrule
    \end{tabular}%
    }
\end{table}

\subsection{Effect of Number of MLP Hidden Dimensions}

We examined the impact of varying MLP hidden dimensions on model performance, with results presented in Table~\ref{tab:mlp_accuracy}. The experiments demonstrate that model performance improves as the hidden dimension size increases from 256 to 768, reaching peak accuracy at 768 dimensions (82.40\%). Further increases to 1024 maintain this performance level, with a marginal improvement to 82.50\% at 1280 dimensions. This suggests that while larger hidden dimensions can capture more complex features, the benefits plateau around 768-1024 dimensions, indicating this range as the optimal choice for balancing model capacity and computational efficiency.

\begin{table}[htbp]
    \centering
    \caption{Accuracy results for different MLP hidden dimensions}
    \label{tab:mlp_accuracy}
    \resizebox{\columnwidth}{!}{%
    \begin{tabular}{ccccccc}
        \toprule
        MLP dimensions &  256 & 512 & 768 & 1024 & 1280 \\
        \midrule
        Accuracy (\%) & 81.00 & 81.60 & 82.40 & 82.40 & 82.50 \\
        \bottomrule
    \end{tabular}%
    }
\end{table}

\subsection{Effect of number of decoders}

We investigated the impact of varying the number of decoders on model performance, as shown in Table~\ref{tab:decoder_accuracy}. The results reveal a clear pattern where using a single decoder yields the lowest accuracy of 78.60\%. A substantial improvement is observed when increasing to 3-4 decoders, both achieving 81.60\%. Further enhancement is achieved with 8 decoders, reaching 82.40\%, which plateaus through 12 decoders. Adding more decoders up to 16 provides only marginal improvement (82.50\%). These results indicate that while multiple decoders are crucial for better feature learning, the performance gains saturate beyond 8 decoders, suggesting this as an optimal choice considering the computational efficiency trade-off.

\begin{table}[htbp]
    \centering
    \caption{Accuracy results for different decoder sizes}
    \resizebox{\columnwidth}{!}{%
    \begin{tabular}{ccccccc}
        \toprule
        No. of decoders & 1 & 3 & 4 & 8 & 12 & 16 \\
        \midrule
        Accuracy (\%) & 78.60 & 81.60 & 81.60 & 82.40 & 82.40 & 82.50 \\
        \bottomrule
    \end{tabular}
    }
    \label{tab:decoder_accuracy}
\end{table}
\section{Conclusion}

In this work, we presented DenoMAE2.0, an enhanced denoising autoencoder that jointly optimizes reconstruction loss and an unmasked patch classification loss to improve robustness and performance. By incorporating a classification head that refines feature learning on unmasked regions, DenoMAE2.0 surpasses its predecessor, DenoMAE, achieving state-of-the-art results across multiple benchmarks. Extensive experiments demonstrate its superior ability to handle noisy data while preserving high classification accuracy. Additionally, we investigate the influence of auxiliary loss weighting and MLP hidden dimensions on performance, offering key insights into the role of hyperparameter tuning in optimizing denoising autoencoders.

\bibliographystyle{IEEEtran}
\bibliography{main}

\clearpage
\section{Diagrams}

\begin{figure*}[hbp]
    \centering
    \includegraphics[width=0.8\linewidth]{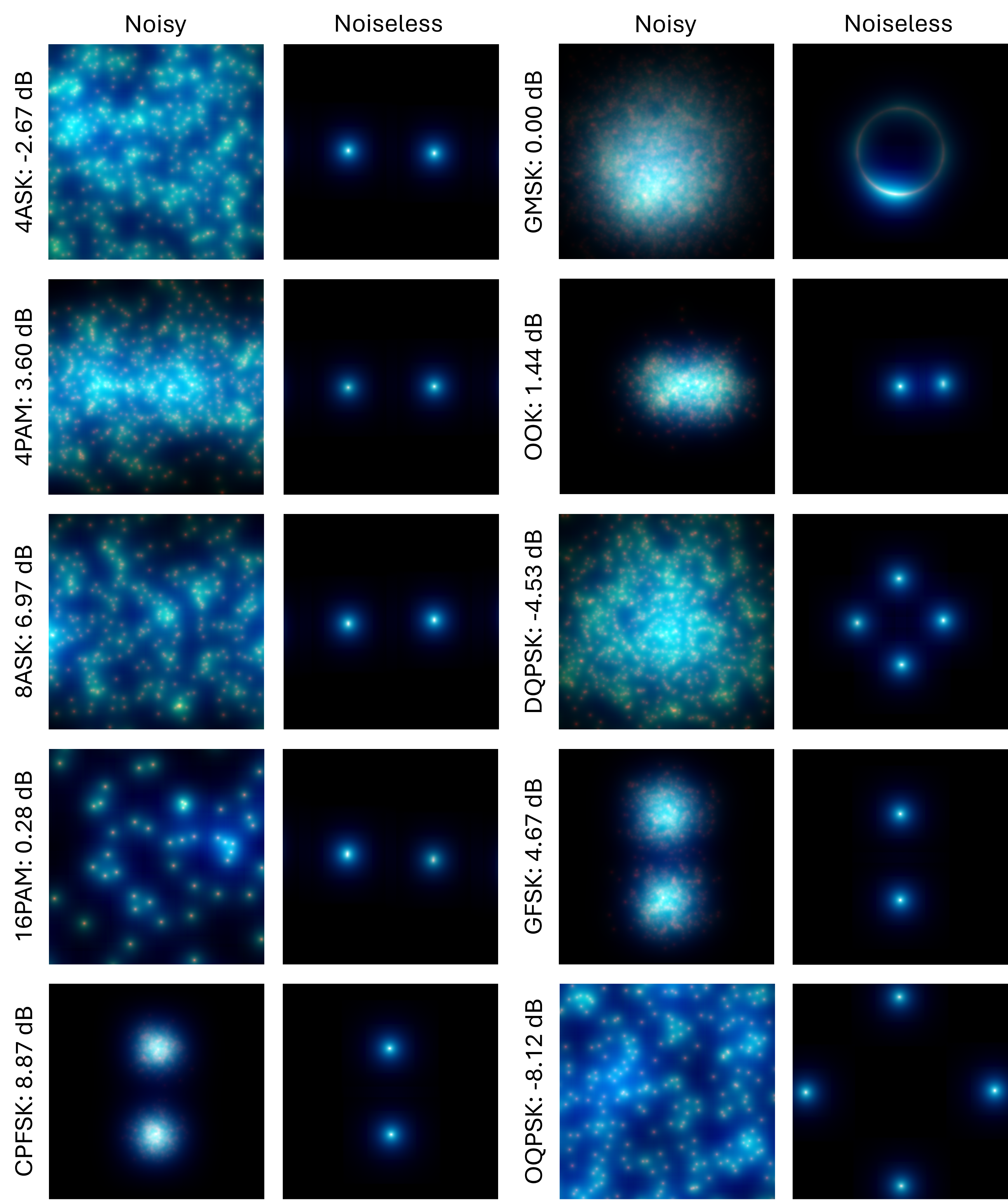}
    \caption{constellation diagram}
    \label{fig:cons}
\end{figure*}

Figure \ref{fig:cons} presents the constellation diagrams derived from the dataset utilized in our research. Each diagram includes a sample from each class along with their corresponding randomly generated noise. The diagrams depict the signal as a two-dimensional scatter plot in the complex plane known as constellation diagrams. Additionally, we provide the noiseless images to illustrate the original signal without noise. The list of classes are provided in Table \ref{tab:cons_classes}.

\begin{table}[h]
    \centering
    \caption{List of classes from our dataset}
    \label{tab:cons_classes}
    \begin{tabular}{|c|c|}
        \hline
        Class Number & Class Name \\ \hline
        1 & 4ASK \\ \hline
        2 & 4PAM \\ \hline
        3 & 8ASK \\ \hline
        4 & 16PAM \\ \hline
        5 & CPFSK \\ \hline
        6 & DQPSK \\ \hline
        7 & GFSK \\ \hline
        8 & GMSK \\ \hline
        9 & OOK \\ \hline
        10 & OQPSK \\ \hline
    \end{tabular}
\end{table}

\begin{figure*}[htbp]
    \centering
    \includegraphics[width=0.8\linewidth]{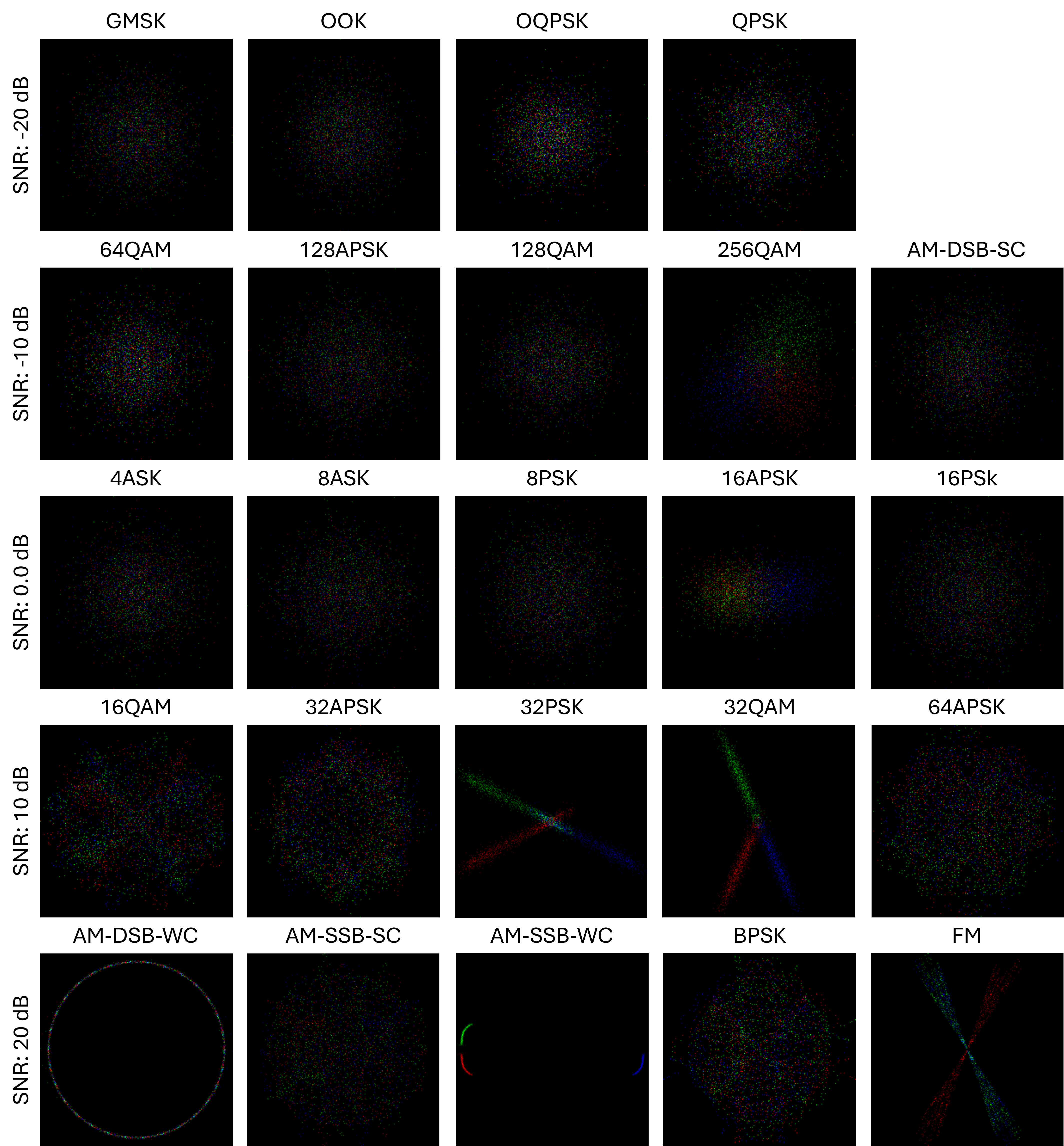}
    \caption{RadioML diagram}
    \label{fig:radio}
\end{figure*}

Figure~\ref{fig:radio} illustrates the RadioML diagram, which is used to visualize the performance of machine learning models in radio signal classification tasks. These constellation diagrams helps in understanding how well the model can distinguish between different types of radio signals. The list of classes are provided in Table~\ref{tab:radioml_classes}.

\begin{table}[h]
    \centering
    \caption{List of classes in the RadioML diagram}
    \begin{tabular}{|c|c|}
        \hline
        Class Number & Class Name \\ \hline
        1 & 4ASK \\ \hline
        2 & 8ASK \\ \hline
        3 & 8PSK \\ \hline
        4 & 16APSK \\ \hline
        5 & 16PSK \\ \hline
        6 & 16QAM \\ \hline
        7 & 32APSK \\ \hline
        8 & 32QAM \\ \hline
        9 & 64APSK \\ \hline
        10 & 64QAM \\ \hline
        11 & 128APSK \\ \hline
        12 & 128QAM \\ \hline
        13 & 256QAM \\ \hline
        14 & AM-DSB-SC \\ \hline
        15 & AM-DSB-WC \\ \hline
        16 & AM-SSB-SC \\ \hline
        17 & AM-SSB-WC \\ \hline
        18 & BPSK \\ \hline
        19 & FM \\ \hline
        20 & GFSK \\ \hline
        21 & OQPSK \\ \hline
        22 & OOK \\ \hline
        23 & OQPSK \\ \hline
        24 & QPSK \\ \hline
    \end{tabular}
    \label{tab:radioml_classes}
\end{table}

\begin{table}[h]
    \centering
    \caption{List of hyperparameters during pretraining}
    \begin{tabular}{|c|c|}
        \hline
        \textbf{Parameter name} & \textbf{Value} \\ \hline
        Training batch size & 64 \\ \hline
        Testing batch size & 10 \\ \hline
        Image size & 224, 224 \\ \hline
        Patch size & 16 \\ \hline
        Mask ratio & 0.75 \\ \hline
        Encoder embedding dimension & 768 \\ \hline
        Decoder embedding dimension & 512 \\ \hline
        Encoder depth & 12 \\ \hline
        Decoder depth & 8 \\ \hline
        Encoder number of heads & 12 \\ \hline
        Decoder number of heads & 8 \\ \hline
        Number of epochs & 100 \\ \hline
        Learning rate & 0.0003 \\ \hline
        Weight decay & 0.05 \\ \hline
        Classification loss weight & 0.1 \\ \hline
        Reconstruction loss weight & 1.0 \\ \hline
    \end{tabular}
    \label{tab:pre_para}
\end{table}

\begin{table}[h]
    \centering
    \caption{List of hyperparameters during pretraining}
    \begin{tabular}{|c|c|}
        \hline
        \textbf{Parameter name} & \textbf{Value} \\ \hline
        Training batch size & 32 \\ \hline
        Testing batch size & 10 \\ \hline
        Image size & 224, 224 \\ \hline
        Patch size & 16 \\ \hline
        Encoder embedding dimension & 768 \\ \hline
        Encoder depth & 12 \\ \hline
        Encoder number of heads & 12 \\ \hline
        Number of epochs & 150 \\ \hline
        Learning rate & 0.0001 \\ \hline
        Weight decay & 0.05 \\ \hline
        Number of classes & 10, 24 \\ \hline
    \end{tabular}
    \label{tab:fine_para}
\end{table}

\end{document}